\documentclass[11pt,a4paper]{article}
\usepackage[hyperref]{acl2021}
\usepackage{times}
\usepackage{latexsym}

\usepackage{caption}
\usepackage{subcaption}

\usepackage{mathtools}
\usepackage{amsfonts}

\DeclareMathOperator*{\argmax}{arg\,max}

\newcommand\argtopk{\operatorname{argtop-k}}
\newcommand\supp{\operatorname{supp}}

\newcommand{\eos}{\ensuremath{\left<\text{eos}\right>\,}}

\newcommand{\pad}{\ensuremath{\left<\text{pad}\right>\,}}

\DeclareMathAlphabet{\mathcal}{OMS}{cmsy}{m}{n}
\usepackage[nameinlink]{cleveref}

\crefformat{equation}{Eq.~\textup{#2#1#3}}
\Crefformat{equation}{Eq.~\textup{#2#1#3}}

\crefformat{figure}{Fig.\textup{#2#1#3}}
\Crefformat{figure}{Fig.~\textup{#2#1#3}}

\crefformat{section}{Sec.\textup{#2#1#3}}
\Crefformat{section}{Sec.~\textup{#2#1#3}}

\usepackage{microtype}

\aclfinalcopy %

\title{Mode recovery in neural autoregressive sequence modeling}

\author{Ilia Kulikov \\
  New York University \\
  \texttt{kulikov@cs.nyu.edu} \\\And
  Sean Welleck \\
  University of Washington \\
  \\\And
  Kyunghyun Cho \\
  New York University \\
  CIFAR Fellow in Learning \\in Machines \& Brains \\
  }

\date{}

\begin{document}
\maketitle
\begin{abstract}
Despite its wide use,
recent studies have revealed unexpected and undesirable properties of neural autoregressive sequence models trained with maximum likelihood, such as an unreasonably high affinity to short sequences after training and to infinitely long sequences at decoding time. We propose to study these phenomena by investigating how the modes, or local maxima, of a distribution are maintained throughout the full {\it learning chain} of the ground-truth, empirical, learned and decoding-induced distributions, via the newly proposed \textit{mode recovery cost}. We design a tractable testbed where we build three types of ground-truth distributions: (1) an LSTM based structured distribution, (2) an unstructured distribution where probability of a sequence does not depend on its content, and (3) a product of these two which we call a semi-structured distribution. Our study reveals both expected and unexpected findings. First, starting with data collection, mode recovery cost strongly relies on the ground-truth distribution and is most costly with the semi-structured distribution.
Second, after learning, mode recovery cost from the ground-truth distribution may increase or decrease compared to data collection, with the largest cost degradation occurring with the semi-structured ground-truth distribution.
Finally, the ability of the decoding-induced distribution to recover modes from the learned distribution is highly impacted by the choices made earlier in the learning chain.
We conclude that future research must consider the entire learning chain in order to fully understand the potentials and perils and to further improve neural autoregressive sequence models.
\end{abstract}

\section{Introduction}

Neural autoregressive sequence modeling has become the standard approach to modeling sequences in a variety of natural language processing applications \citep{aharoni2019massively, brown2020language, roller2020recipes}. 
In this modeling paradigm, the probability of a sequence is decomposed into the product of the conditional probability of each token given the previous tokens. 
Each conditional probability is modeled by a shared neural network, typically implemented as a recurrent neural network \citep{Hochreiter1997LongSM} or a transformer \citep{vaswani2017attention}. 

Despite its success, recent studies have identified peculiarities in neural autoregressive sequence models. 
\citet{lee2018hallucinations} identify \textit{hallucinations} in neural machine translation, in which a well-trained model suddenly generates a nonsense translation when a rare token is artificially introduced to a source sentence. 
\citet{stahlberg2019nmt} observe that a vast portion of probability mass is concentrated on the \textit{empty sequence} in neural machine translation, although the models they studied were never presented with empty sequences during training. \citet{holtzman2019curious} report that large-scale language models often produce pathological sequences with many \textit{n-gram repetitions}, at a rate which far exceeds that of the training data. 
\citet{welleck2020consistency} show that neural language models can generate \textit{infinite-length sequences} despite being trained on only finite sequences.

A common theme underlying these findings is that well-trained models can assign unreasonably high probabilities to sequences that are dissimilar to any sequence from the training set. 
In particular, the \textit{modes} of the model's distribution appear to be undesired, implying that the model failed to \textit{recover} the modes of the empirical distribution, which we term \textit{mode recovery degradation}.
The situation is further complicated by the fact that we only approximate the model's modes with a decoding algorithm, so it is unclear whether the decoding algorithm, the model, or even the data collection is at fault.

In this paper, we isolate and study mode recovery degradation by characterizing each stage of neural sequence modeling as inducing a new sequence distribution, then directly analyzing each distribution's modes.
With this approach, we diagnose at what stage, and to what extent, sequences receive unreasonably high probabilities.
To do so, we first define a \textit{learning chain} that consists of the \textit{ground-truth} distribution, the \textit{empirical} distribution induced by data collection, the \textit{learned} distribution, and the \textit{decoding-induced} distribution.  
We then quantify the extent to which the most probable sequences under each distribution match the most probable sequences under the ground-truth distribution by defining a \textit{mode recovery cost},
which measures how expensive it is for a later distribution to recover the most probable sequences of an earlier distribution in the chain. 

In summary, we find that mode recovery cost is non-trivial at each part of the neural autoregressive learning pipeline. The pattern of how mode recovery changes heavily depends on the properties of the ground-truth distribution. In particular, when the ground-truth distribution is parameterized as a product of highly structured distribution based on LSTM neural network and unstructured distribution where the probability of every sequence is sampled independently from all the others, its modes are more costly to recover. Furthermore, the ability of a decoding algorithm to recover modes is also dependent upon all choices made earlier in the chain including the underlying ground-truth distribution, even in the case of modes of the learned distribution. These observations make  a meaningful step towards better understanding of mode degradation in neural autoregressive sequence modeling.

\section{Neural autoregressive sequence modeling}
We consider the problem of modeling a distribution $p^{*}(\mathbf{s})$ over variable-length, discrete sequences $\mathbf{s}$.
Formally, $\mathbf{s} \in \Sigma^l$, where $l \in \left\{1, 2, \ldots, L \right\}$,  $\Sigma$ is a finite set of tokens, and $\Omega \subset \bigcup_{l=1}^{L} \Sigma^l$ denotes the space of all possible sequences.
Every sequence $\mathbf{s} \in \Omega$ ends with a special token $\eos \in \Sigma$ which  only appears at the end of each sequence.

In neural autoregressive sequence modeling, we model the distribution $p^*(\mathbf{s})$ as
$%
    p_\theta(\mathbf{s}) = \prod_{t=1}^{|\mathbf{s}|} p_\theta(s_t | s_{<t})$,
with each conditional distribution parameterized by a shared neural network.

\paragraph{Maximum likelihood.}
To learn the model, we use maximum likelihood estimation (MLE), which trains the model $p_{\theta}$ to maximize the log-likelihood of a set of training sequences $D=\left\{ \mathbf{s}^1, \ldots, \mathbf{s}^N \right\}$:
\begin{align}
\label{eqn:mle}
    \argmax_{\theta} \frac{1}{N} \sum_{n=1}^N \sum_{t=1}^{L^n} \log p_\theta(s^n_t | s^n_{<t}).
\end{align}

\paragraph{Approximate decoding.}\label{par:approximate_decoding}

Given a trained model, we obtain a set of highly probable sequences.
In practice, this problem is often intractable due to the size of $\Omega$, which grows exponentially in sequence length.
As a result, we resort to approximating the optimization problem using a decoding algorithm that returns a set of $k$ sequences $\mathcal{F}(p_{\theta}; \gamma)$,
where $\mathcal{F}$ denotes the decoding algorithm, and $\gamma$ denotes its hyper-parameters.
Concretely, we consider two decoding approaches: a \textit{deterministic} decoding algorithm  that produces a set of sequences using beam search with beam-width $k$, and a \textit{stochastic} decoding algorithm that forms a set of sequences using ancestral sampling until $k$ unique sequences are obtained.\footnote{Ancestral sampling recursively samples $s_t\sim p_{\theta}(s_t\vert s_{<t})$. } We refer readers to \citet{welleck2020consistency} for detailed descriptions of those decoding algorithms.

\paragraph{Learning chain.}
The neural autoregressive sequence modeling approach consists of four probability distributions, which together form a \textit{learning chain.}
The first distribution is the \textit{ground-truth distribution} $p^*(\mathbf{s})$.
This distribution is almost always unknown and is assumed to be highly complicated. Second, the dataset used in maximum likelihood (\cref{eqn:mle}) determines an {\it empirical distribution}, 
\begin{align}
    \label{eqn:empirical_prob}
    p_{\mathrm{emp}}(\mathbf{s}) 
    =
    \frac{1}{|D|}
    \sum_{\mathbf{s}' \in D}
    \mathbb{I}(\mathbf{s} = \mathbf{s}'), 
\end{align}
where $D$ is a set of sequences drawn from the ground-truth distribution $p^*$ and $\mathbb{I}$ is the indicator function.
The third distribution is the {\it learned distribution} $p_{\mathrm{model}}$ captured by a neural autoregressive model trained on $D$. 

Finally, we introduce the \textit{decoding-induced distribution} $p_\mathcal{F}$, which allows us to compare the set of probable sequences obtained with a decoding algorithm $\mathcal{F}$ 
against highly probable sequences in the ground-truth, empirical, and learned distributions. 
Specifically, we turn this set into the distribution
\begin{align}
\label{eqn:qf}
    p_{\mathcal{F}}(\mathbf{s}) &= \begin{cases} 
        \frac{1}{Z}p_{\theta}(\mathbf{s}) & \mathbf{s}\in \mathcal{F}(p_{\theta}; \gamma), \\
        0 & \mathbf{s}\not\in \mathcal{F}(p_{\theta}; \gamma),
        \end{cases}
\end{align}
where $Z=\sum_{\mathbf{s}'\in \mathcal{F}(p_{\theta}; \gamma)}p_{\theta}(\mathbf{s}')$.
Each sequence is weighted according to the model's probability, which reflects the practice of ordering and sampling beam search candidates by their probabilities.

There is a natural order of dependencies among these four distributions in the learning chain, ${p^* {\succ}_{\text{data collection}}~~p_\mathrm{emp} 
{\succ}_{\text{learning}}~~p_\mathrm{model} 
{\succ}_{\text{decoding}}~~p_\mathcal{F}}$.
We are interested in how a distribution in the later part of the chain recovers the highly probable sequences of an earlier distribution.
To study this, we next introduce the notion of \textit{mode recovery}.

\section{Mode recovery}
\label{sec:mode_recovery}
\paragraph{Mode sets}

We define a {\it $k$-mode set} as a set of top-$k$ sequences under a 
given distribution:
\begin{align*}
    \mathcal{S}_k(p) = 
    \argtopk_{\mathbf{s} \in \Omega} p(\mathbf{s}).
\end{align*}
$\argtopk$ selects all the elements within $\Omega$ whose probabilities $p(\mathbf{s})$ are {\it greater than} the probability assigned to the $(k+1)$-st most likely sequence, which could result in fewer than $k$ sequences. This is due to potentially having multiple sequences of the same probability.

\paragraph{Mode recovery cost.}

We characterize the recovery of the modes of the distribution $p$ by the distribution $q$ as the cost required to recover the $k$-mode set $\mathcal{S}_{k}(p)$ using the distribution $q$. 
That is, how many likely sequences under $q$ must be considered to recover all the sequences in the $k$-mode set of $p$. 

Formally, given a pair of distributions $p$ and $q$, we define the $k$-mode recovery cost from $p$ to $q$ as
\begin{align}
\label{eqn:kcost}
    \mathcal{O}_{k}(p \| q)
    =
    \min
    \left\{
    k' 
    \Big|
    \mathcal{S}_{k}(p)
    \subseteq
    \mathcal{S}_{k'}(q)
    \right\}.
\end{align}
The cost is minimized ($=|\mathcal{S}_{k}(p)|$) when the $k$-mode set of $q$ perfectly overlaps with that of $p$. 
The cost increases toward $|\Omega|$ as the number of modes from $q$ that must be considered to include the $k$-mode set from $p$ increases. 
The cost is maximized (=$|\Omega|$) when the top-$k$ set $\mathcal{S}_{k}(p)$ of $p$ is not a subset of the support of the distribution $q$.

\paragraph{The limited support of $q$.} 

As mentioned earlier, the mode recovery cost $\mathcal{O}_k (p\|q)$ is ill-defined when the support of the distribution $q$, $\supp(q)$, is not a super-set of the $k$-mode set of the distribution $p$ . In this situation, we say that the distribution $q$ fails to recover modes from the $k$-mode set of the distribution $p$. In particular, this happens with decoding-induced distributions because of their limited support, which is equal to the size of the candidate set of sequences $\mathcal{F}(p_\theta, \gamma)$.

We introduce the $k$-mode set overlap
    \mbox{$\mathcal{I}_k(p\|q) = | \mathcal{S}_k(p) \cap \supp(q) |$},
which equals the size of the intersection between the $k$-mode set of the distribution $p$ and the support of the distribution $q$. The $k$-mode set overlap is maximized and equals $|\mathcal{S}_k(p)|$ when the mode recovery is successful. We call it a recovery failure whenever the overlap is smaller than $|\mathcal{S}_k(p)|$. We use $k$-mode set overlap only when mode recovery fails, because it is not able to detect if the modes from the corresponding $k$-mode set have high probability under the induced distribution.

\section{Why do we study mode recovery?}

The recent success of neural sequence modeling has operated on the assumption that we can find sequences that are reasonably similar to training sequences by fitting a neural autoregressive model to maximize the log-probabilities of the training sequences (maximum-likelihood learning) and searching for the most likely sequences under the trained model (maximum a posteriori inference).
However, 
recent studies suggest that the most likely sequences may not resemble training sequences at all.
For instance, the learning stage can yield a distribution $p_{\text{model}}$ which places high probability on empty \citep{stahlberg2019nmt} or repetitive \citep{holtzman2019curious} sequences, while the decoding stage can yield a distribution $p_{\mathcal{F}}$ which places non-zero mass on infinite-length sequences \citep{welleck2020consistency}.

As a result, various workarounds have been proposed in the form of alternative learning or decoding algorithms \citep{andor2016globally,sountsov2016length,murray2018correcting,welleck2020unlikelihood,welleck2020mleguided,martins2020sparse,deng2020Residual,basu2021mirostat, shi2020neural}.
A particularly relevant work by \citet{eikema2020map} argues that the modes of neural sequence models are \textit{inadequate} and thus we must discard maximum-a-posteriori inference altogether.
Rather than advocating for a particular solution, we instead seek an understanding of \textit{why} the conventional approach displays these peculiar behaviors.
While we do not claim to provide a full explanation, the first step is developing a way of quantifying the problem, then localizing it.
To this end, we develop the mode recovery cost and measure it along the learning chain $p^* {\succ} p_\mathrm{emp} 
{\succ} p_\mathrm{model} 
{\succ} p_\mathcal{F}$.
This focus on modes departs from the conventional focus on evaluating the full distribution with 
a probabilistic divergence.

\paragraph{Mode recovery vs. probabilistic divergence.}

Mode recovery is related to but distinct from a probabilistic divergence. Often a probabilistic divergence is designed to consider the full support of one of two distributions between which the divergence is computed.  For each point within this support, a probabilistic divergence considers the ratio, or difference, between the actual probabilities/densities assigned by the two distributions. 
For instance, the KL divergence $\mathrm{KL}(p \| q)$ computes
$
    \sum_{x \sim p} p(x)\log p(x)\big/q(x).
$
Another example is the total variation (TV) distance, which is equivalent to $\sum_{\omega \in \Omega} |p(\omega) - q(\omega)|/ 2$ when the sample set $\Omega$ is finite. The TV distance considers the entire sample set and computes the cumulative absolute difference between the probabilities assigned to each event by two distributions. 

We find mode recovery more interesting than probabilistic divergence in this paper, because our goal is to check whether a decision rule, that is to (approximately) choose the most likely sequence based on an available distribution, changes as we follow the chain of induced distributions. 
Furthermore, we are not interested in how precisely unlikely sequences are modeled and what probabilities they are being assigned. We thus fully focus on mode recovery in this paper.

\section{A testbed for evaluating mode recovery}
\label{sec:testbed}

It is intractable to measure mode recovery cost (\cref{eqn:kcost}) on real-world datasets that are popular in neural sequence modeling, e.g. wikitext-103 \citep{merity2016pointer}  given the exponential growth of the sequence space with sequence length. For example, the training part of Wikitext-103 consists of ~$28$k sequences with ~$3.5$k tokens, each drawn from a vocabulary of $267$k tokens. 
Furthermore, these datasets do not provide access to the ground-truth distribution, which prevents us from computing any recovery cost involving $p^*$.

In order to allow exact computations of mode recovery cost, we design a controllable testbed. This testbed consists of (1) the ground-truth distribution, which permits explicit control over the structuredness, (2) the data collection step, which controls the complexity of the empirical distribution, (3) the learning step, which allows us to induce the learned distribution with neural autoregressive models, and (4) the decoding step, where the decoding algorithm induces the approximation of the learned distribution. In the rest of this section we describe each distribution in detail.

We set the size of the sequence space of the testbed so that all computations are feasible. We limit the vocabulary size $|\Sigma|$ to 7 tokens and use a maximum sequence length $L$ of 10 tokens.
This results in a sequence space size $|\Omega|$ of around 12 million sequences.

\paragraph{Ground-truth distribution.}
\phantomsection
\label{ssec:true_distribution}
We define each ground-truth distribution as a product of two components: 
\begin{align*}
    p^*_{\alpha}(\mathbf{s}) &\propto p_\theta(\mathbf{s})^\alpha p(\mathbf{s};\mu, \sigma)^{(1-\alpha)},
\end{align*}
where $p_\theta(\mathbf{s})$ is an autoregressive distribution with parameters $\theta$. The probability $p(\mathbf{s};\mu, \sigma)$ is constructed by
$    p(\mathbf{s};\mu, \sigma) \propto \exp(x(\mathbf{s}))$,
where $x(\mathbf{s}) \sim \text{Laplace}(\mu, \sigma)$ is a fixed random sample for each $\mathbf{s}$, and $\alpha \in [0, 1]$. 

We implement $p_\theta$ using a randomly initialized LSTM neural network, with two layers and $512$ LSTM units in every layer. We build $p(\mathbf{s};\mu, \sigma)$ 
with $\mu=0.0$ and $\sigma=1.0$.

We build the ground-truth distribution to reflect some properties of real data.
First, real data has strong statistical dependencies among the tokens within each sequence. We induce these dependencies by assuming that each sequence is produced from left to right by generating each token conditioned on the previously generated sub-sequences of tokens. We implement this procedure using the LSTM neural network.

Second, there exist exceptional sequences in real data which receive high probability even though those sequences do not reflect statistical dependencies mentioned above.
We build another distribution component in order to introduce exceptions in a way that there are no statistical dependencies in the given sequence. We use independent samples from a Laplace distribution 
as unnormalized probabilities of every sequence from the sequence space $\Omega$. We thus ensure that there are no statistical dependencies among the tokens under this unstructured distribution.

We thus construct the product of two distributions described above so that it exhibits structured and unstructured aspects of the generating process. The mixing coefficient $\alpha$ allows us to interpolate between the heavily structured to heavily unstructured ground-truth distributions. We call it 
\textit{semi-structured} 
when $0 < \alpha < 1$.

\begin{figure}
    \centering
    \includegraphics[width=1.\linewidth]{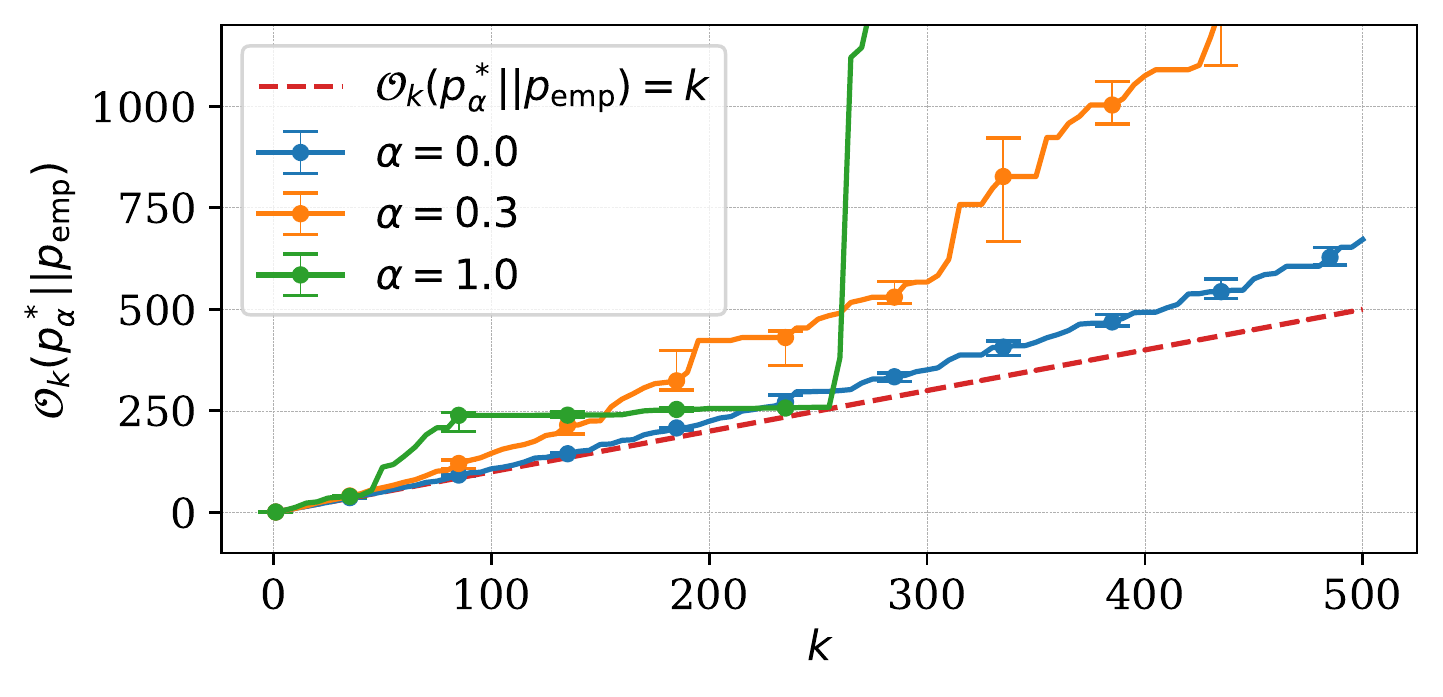}
    \caption{Mode recovery cost of the empirical distribution from the ground-truth distribution as a function of $k$ while $N_\text{train} = 5\times 10^5$. %
    }
    \label{fig:cost_true_emp_ts5e5}
\end{figure}

\paragraph{Empirical distribution.}
We create each empirical distribution $p_\text{emp}$ (\cref{eqn:empirical_prob}) by drawing samples with replacement from the ground-truth distribution.
We sample a training multi-set and a validation multi-set, then form the empirical distribution with their union .
We denote the size of the training dataset as $N_{\text{train}}$, and set the size of the validation set to $.05 \times N_{\text{train}}$.

\paragraph{Learned distribution.}
We obtain each learned distribution $p_\text{model}$ by training an LSTM model on the training dataset $D_\text{train}$ using maximum likelihood (\cref{eqn:mle}). 
We vary the complexity of the learned distribution using the number of LSTM units
of every layer of the LSTM neural network from the set $N_\text{model hs} \in \{128, 512\}$. Variable-length sequences are padded with a \pad token in order to form equal-length batches of $5120$ sequences. 
We use the Adam optimizer \citep{kingma2014adam} with a learning rate of $10^{-4}$. 
We compute validation loss every $5\times10^2$ steps, and apply early stopping with a patience of $5$ validation rounds based on increasing validation loss. 
We train the model for up to $2\times 10^4$ steps. 
After training, the checkpoint with the lowest validation loss is selected to parameterize the learned distribution $p_\text{model}$.

\paragraph{Decoding-induced distribution.}
We form decoding-induced distributions (\cref{eqn:qf}) using beam search and ancestral sampling.
For beam search, we set $N_\text{beam} = 500$.
For ancestral sampling, we sample sequences and discard duplicates until a given number of unique sequences, $N_\text{anc} = 500$, are obtained.

\paragraph{Randomness.}
To account for randomness that occurs when initializing the ground-truth distribution, sampling the empirical distribution, and using ancestral sampling during decoding, we run each configuration of the learning chain (i.e. ground-truth, empirical, learned, and decoding-induced distributions) with $10$ different random seeds, and report the median and $25$-th and $75$-th quantiles, if available, of each evaluation metric.

\section{Mode recovery in the learning chain}

We use our testbed
to empirically study mode recovery degradation by measuring mode recovery cost in the data collection,
learning,
and decoding,
stages of the learning chain. 
We use $k \leq 500$.

\paragraph{Data collection: recovering  ground-truth modes with the empirical distribution.}
\phantomsection
\label{ssec:expr-data}

\begin{figure}
    \centering
    \includegraphics[width=1.\linewidth]{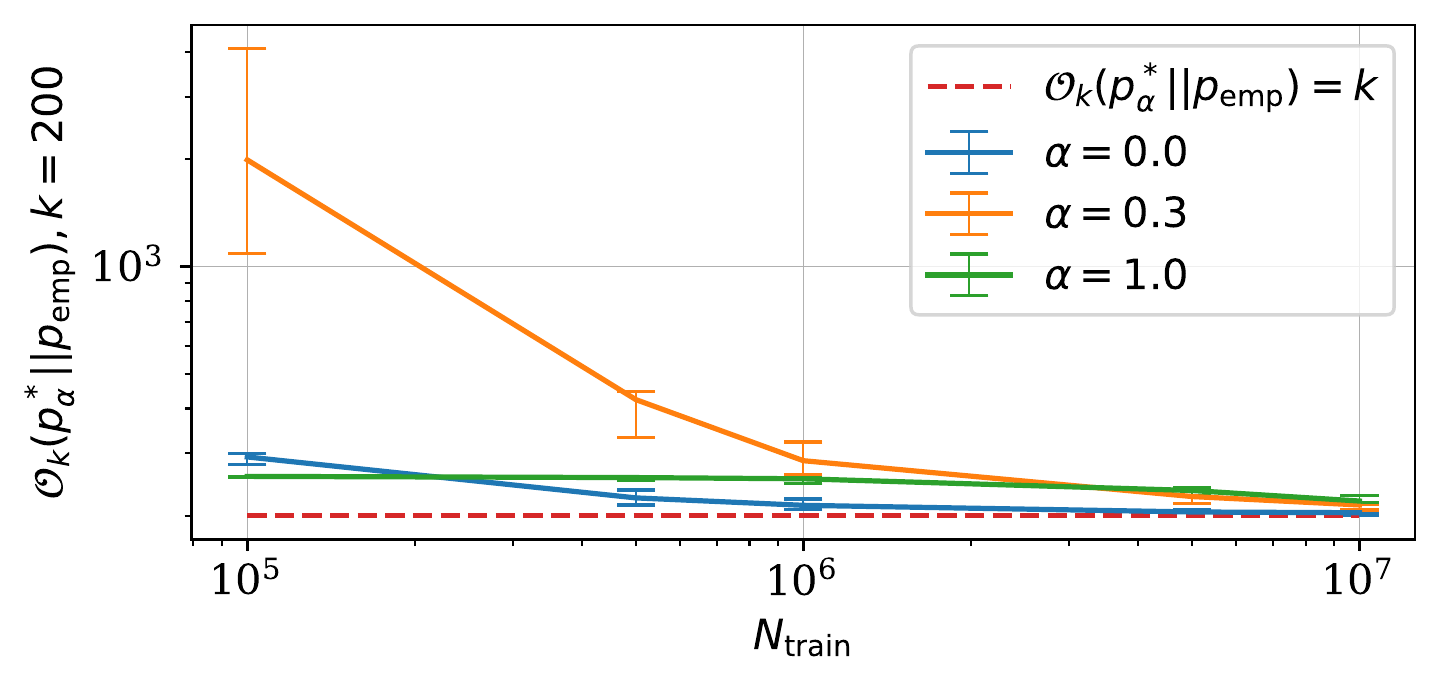}
    \caption{Mode recovery cost of the empirical distribution from the ground-truth distribution as a function of $N_\text{train}$ while $k = 200$. 
    }
    \label{fig:cost_true_emp_allts}
\end{figure}

We start by asking: \textit{does mode degradation happen during data collection?}
We fix $N_\text{train} = 5\times10^5$ and compute mode recovery cost from the ground-truth distribution with the empirical distribution for the range of $k \leq 500$ presented in \cref{fig:cost_true_emp_ts5e5} using three configurations of ground-truth distributions.
 It shows that mode recovery cost grows as $k$ increases. Furthermore, we observe different patterns of mode recovery cost given each choice of the ground-truth distribution.

We observe distinct patterns of mode recovery with either structured ($\alpha=1.0$) and unstructured ($\alpha=0.0$) ground-truth distributions. We found that the structured ground-truth distribution assigns higher probabilities to shorter sequences because of LSTM neural network and autoregressive factorization. This implies that sequences which are sorted w.r.t. their probabilities are also sorted w.r.t. their lengths. Because of this property the empirical distribution can recover modes from the structured ground-truth distribution almost perfectly for particular $k$.
In the case of the unstructured ground-truth distribution mode recovery cost is lower compared to other cases. This ground-truth distribution has no statistical dependencies within modes which makes it less interesting to us due to the lack of similarity with real data.

Finally, in the case of the semi-structured ground-truth distribution ($\alpha=0.3$) the cost of recovering its modes grows increasingly as $k$ increases. In other words, empirical distributions recover modes from ground-truth distributions less effectively when latter exhibit statistical dependencies as well as many exceptional sequence probabilities. 

Now we focus on the influence of the training set size $N_\text{train}$ on mode recovery during data collection. We fix $k=200$ and compute mode recovery cost from the ground-truth distribution using the empirical distribution when $N_\text{train} \in \{10^5, 5 \times 10^5, 10^6, 5 \times 10^6, 10^7\}$, shown in \cref{fig:cost_true_emp_allts}. Mode recovery cost naturally decreases as we increase the number of training instances as seen on the right-most side of \cref{fig:cost_true_emp_allts}. The left-most side is more interesting to us because it corresponds to values of $N_\text{train}$ that reflect real world problems. For instance, in the case of $N_\text{train}=10^5$ it is significantly more costly to recover modes from the semi-structured ground-truth distribution compared to both structured and unstructured variants. We thus conclude that mode recovery degradation happens already during data collection, and that parameterization of ground-truth distributions impacts mode recovery cost.

\paragraph{Learning: recovering modes with the learned distribution.}
\label{ssec:expr-learning}

\begin{figure}
    \centering
    \includegraphics[width=1.\linewidth]{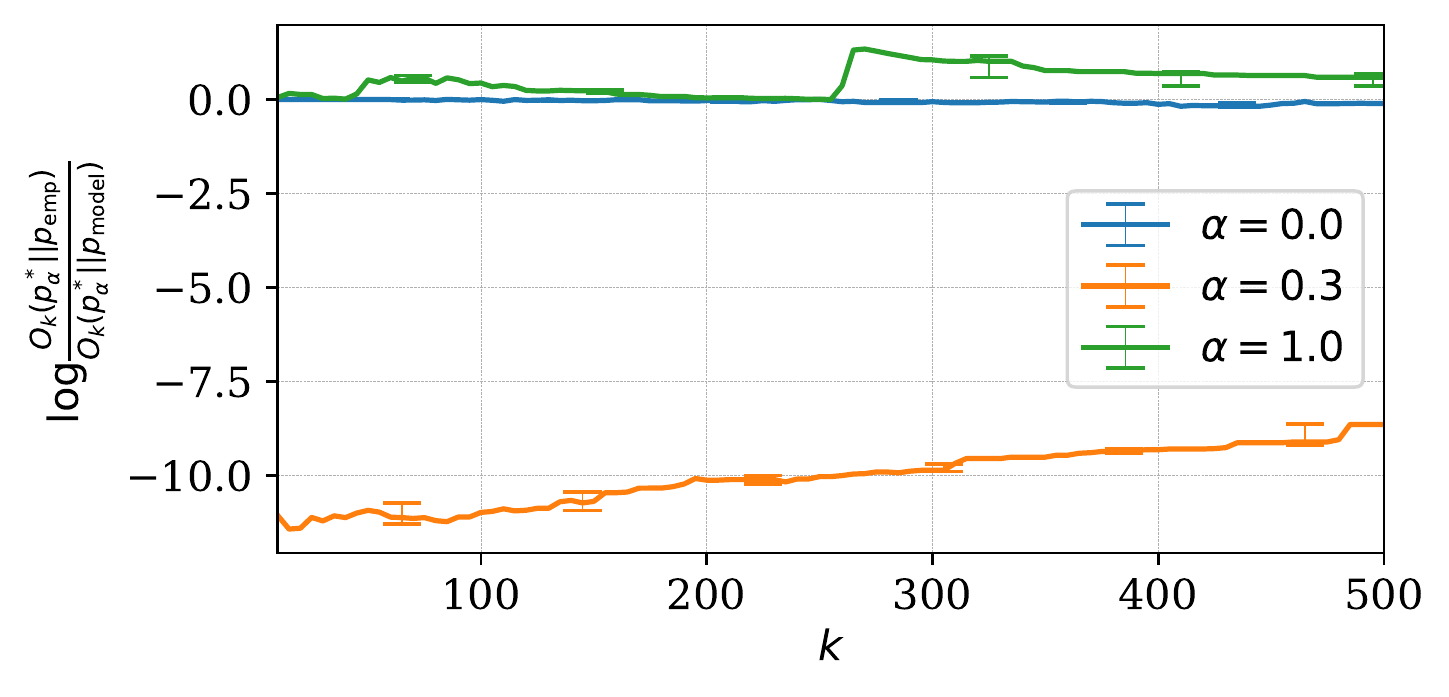}
    \caption{Mode recovery cost reduction log-rate between empirical and learned distributions from the ground-truth distribution as a function of $k$ while $N_\text{train} = 5\times 10^5$. 
    }
    \label{fig:cost_log_ratio_emp_model}
\end{figure}

The next stage in the chain is learning, $~p_\mathrm{emp} 
{\succ}_{\text{learning}}~~p_\mathrm{model}$, in which we train a model using a training dataset with the expectation that the model will match the ground-truth distribution.
Our experiments center on the question: \textit{how does mode recovery degradation in the learning stage compare to that of the data collection stage?}
For instance, we anticipate that the learned model will have a mode recovery cost that is at least as bad as that of the empirical distribution.

We measure the \textit{mode recovery cost reduction log-rate} from empirical to learned distributions, ${\log \frac{\mathcal{O}_{k}(p^*_{\alpha}\|p_\text{emp})}{\mathcal{O}_{k}(p^*_{\alpha}\|p_\text{model})}}$.
\cref{fig:cost_log_ratio_emp_model} shows the reduction log-rate as a function of $k$ with fixed $N_\text{train} = 5 \times 10^5$, for three different ground-truth distributions. We observe three different cases, with a clear dependency on what kind of data was used during learning. 

Learning with data coming from the unstructured ground-truth distribution ($\alpha=0.0$) results in mode recovery cost reduction log-rate being close to zero. This implies that the underlying LSTM model is able to memorize the unstructured data points coming from the empirical distribution, but it can not recover any other modes from the ground-truth distribution.

With the structured ground-truth distribution ($\alpha=1.0$), we observe positive log-rate for some values of $k$. This means that the learned distribution is able to recover modes of the ground-truth distribution at a lower cost than the empirical distribution does.
Similarly to data collection stage, this largely happens due to the property of LSTM to put high probabilities on short sequences.
The learned distribution's ability in mode recovery goes above that of the empirical distribution when there is a match between the parameterization of models behind the ground-truth distribution and the learned distribution. 

In the case of the semi-structured ground-truth distribution ($\alpha=0.3$), the learned distribution has severe mode recovery degradation even with smaller values of $k$ (left-most side of \cref{fig:cost_log_ratio_emp_model}). The model is 
unable to perfectly learn an underlying dataset which has a few statistical exceptions within it.
 
\begin{figure}
    \centering
    \includegraphics[width=1.\linewidth]{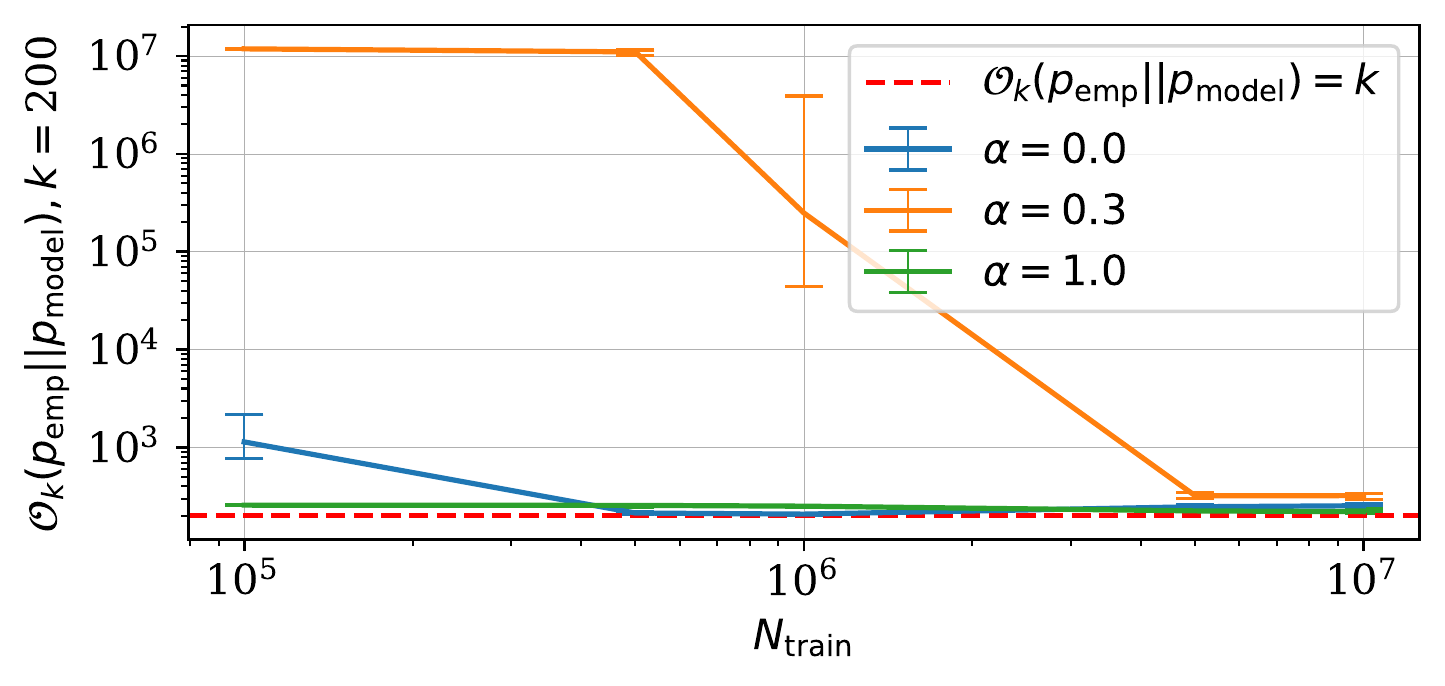}
    \caption{Mode recovery cost of the learned distribution from the empirical distribution as a function of $N_\text{train}$ while $k = 200$. 
    }
    \label{fig:cost_emp_model_allts}
\end{figure}

In addition to our observations about recovering modes from ground-truth distributions, \cref{fig:cost_emp_model_allts} shows at what cost modes of each empirical distribution are recovered by the learned distribution as a function of $N_\text{train}$. 
The learned distribution recovers modes of the empirical distribution with the highest cost when the latter was induced using the semi-structured ground-truth distribution.
Mode recovery cost of all empirical distributions naturally decreases as number of training instances $N_\text{train}$ becomes unrealistically high. We conjecture that the combination of sequences with statistical dependencies and sequences which do not share any statistical dependencies in the dataset makes the learned distribution struggling at mode recovery from both ground-truth and empirical distributions.

\begin{figure*}
\centering
\begin{subfigure}{.5\textwidth}
    \includegraphics[width=1.\linewidth]{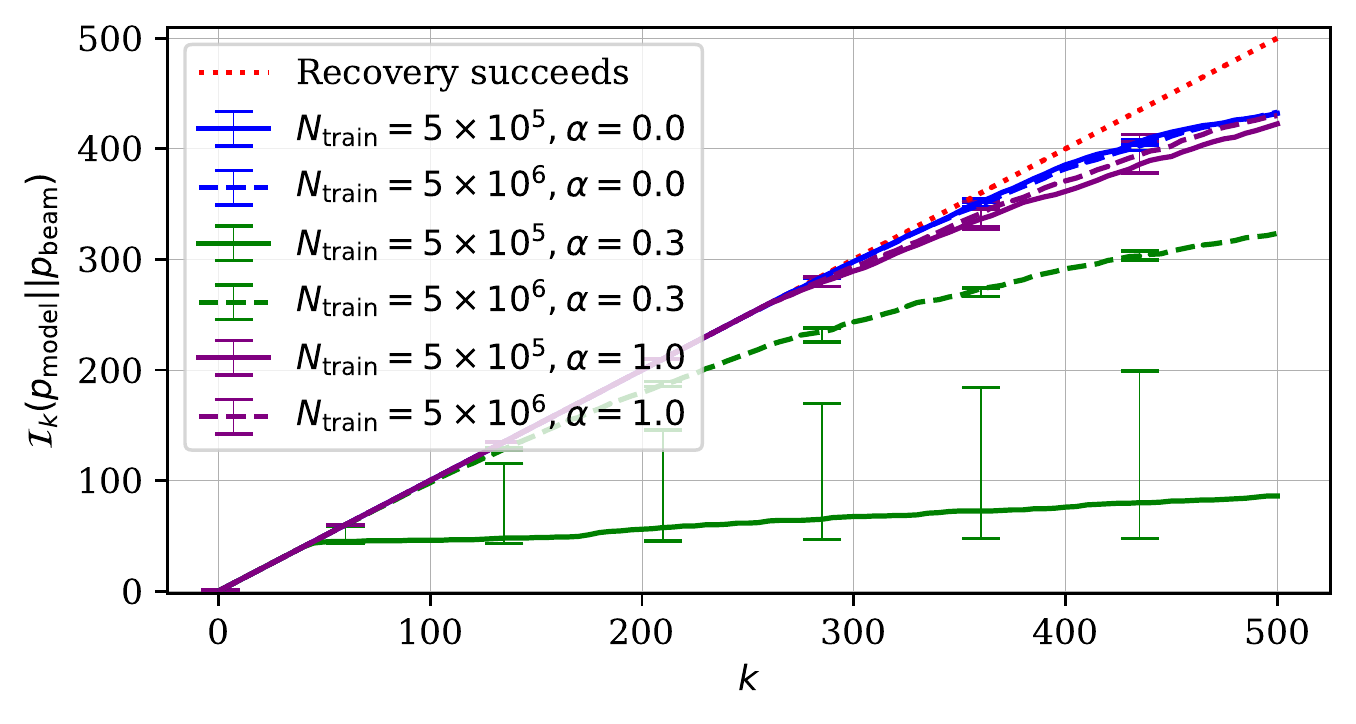}
\end{subfigure}%
\begin{subfigure}{.5\textwidth}
    \includegraphics[width=1.\linewidth]{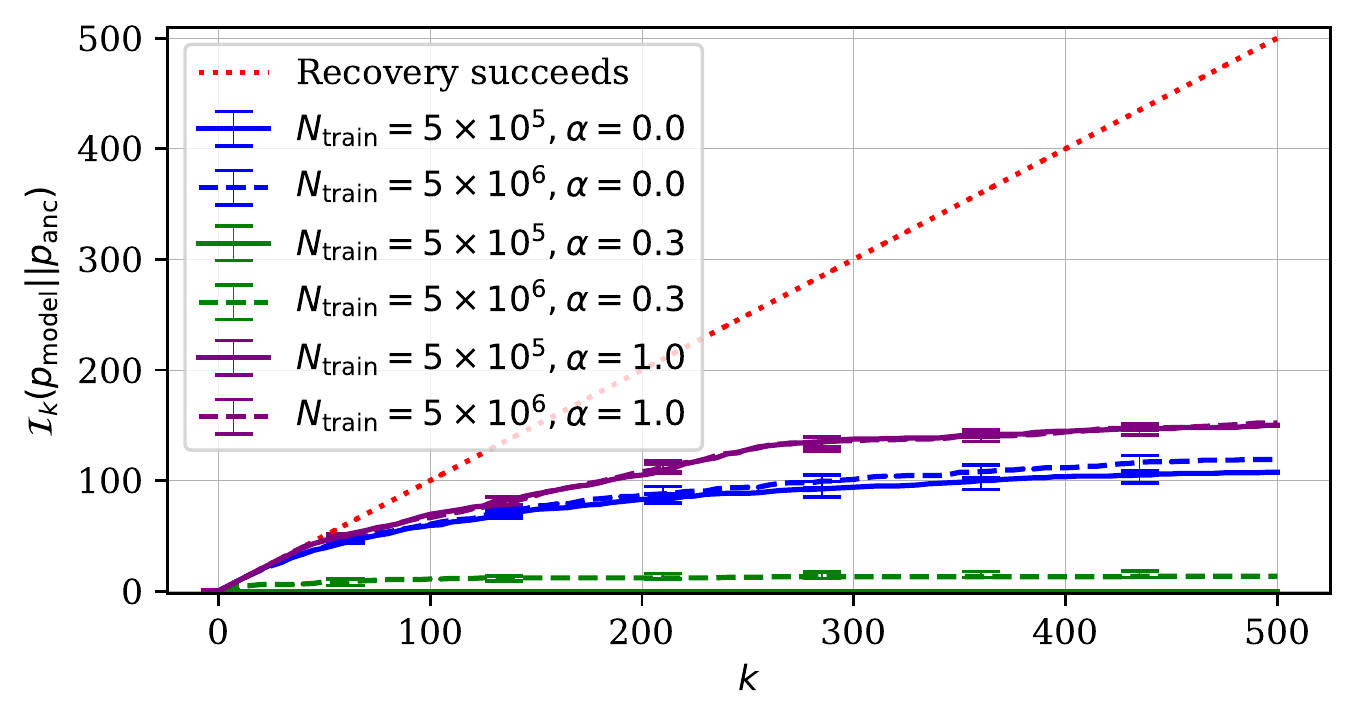}
\end{subfigure}
\caption{$k$-mode set overlap between the learned distribution and the decoding-induced distribution as a function of $k$. 
Choices made earlier in the learning chain (including ground-truth distribution, data collection and learning) affect the degree to which the decoding-induced distribution fails to recover modes from the learned distribution.
}
\label{fig:kmodeoverlap_model}
\end{figure*}

We conclude that properties of ground-truth distributions have direct impacts on the ability of the learned distributions to recover modes from ground-truth and empirical distributions. Learning struggles to capture all patterns from the underlying distributions when the latter exhibit exceptions in statistical dependencies within data points.

\paragraph{Decoding: recovering modes with the decoding-induced distribution.}
\label{ssec:expr-decoding}

The final stage in the learning chain is decoding, $p_\mathrm{model} 
{\succ}_{\text{decoding}}~~p_\mathcal{F}$, in which we use a decoding algorithm $\mathcal{F}$ to obtain highly-probable sequences.
We study both a \textit{deterministic} decoding algorithm, implemented using beam search, and a \textit{stochastic} decoding algorithm, implemented using ancestral sampling.
Our experiments are centered on two questions: (1) \textit{how do the choices made earlier in the learning chain affect the decoding behavior?} and (2) \textit{how is this behavior affected by the choice of the decoding algorithm?}

We consider six different datasets that we train models on, each of which is a combination of the ground-truth distribution where $\alpha \in \{0.0, 0.3, 1.0\}$, and the number of training points $N_\text{train} \in \{5\times10^5, 5\times10^6\}$. 
Our previous analysis revealed each of those datasets leads to a substantially different ability of the learned distributions to recover modes from earlier distributions along the learning chain.
We set $N_\text{model hs}$ to be equal to $512$.
Our choice of decoding algorithms results in decoding-induced distributions with a limited support. Hence the induced distribution $p_\mathcal{F}$ often fails to recover modes of distributions from the earlier stage of the chain especially as $k$ increases. As we described in \Cref{sec:mode_recovery}, we use the $k$-mode set overlap $\mathcal{I}_k(\cdot\|p_\mathcal{F})$ to examine the degree to which a given decoding algorithm $\mathcal{F}$ fails at mode recovery.

\begin{figure*}
\centering
\begin{subfigure}{.5\textwidth}
    \includegraphics[width=1.\linewidth]{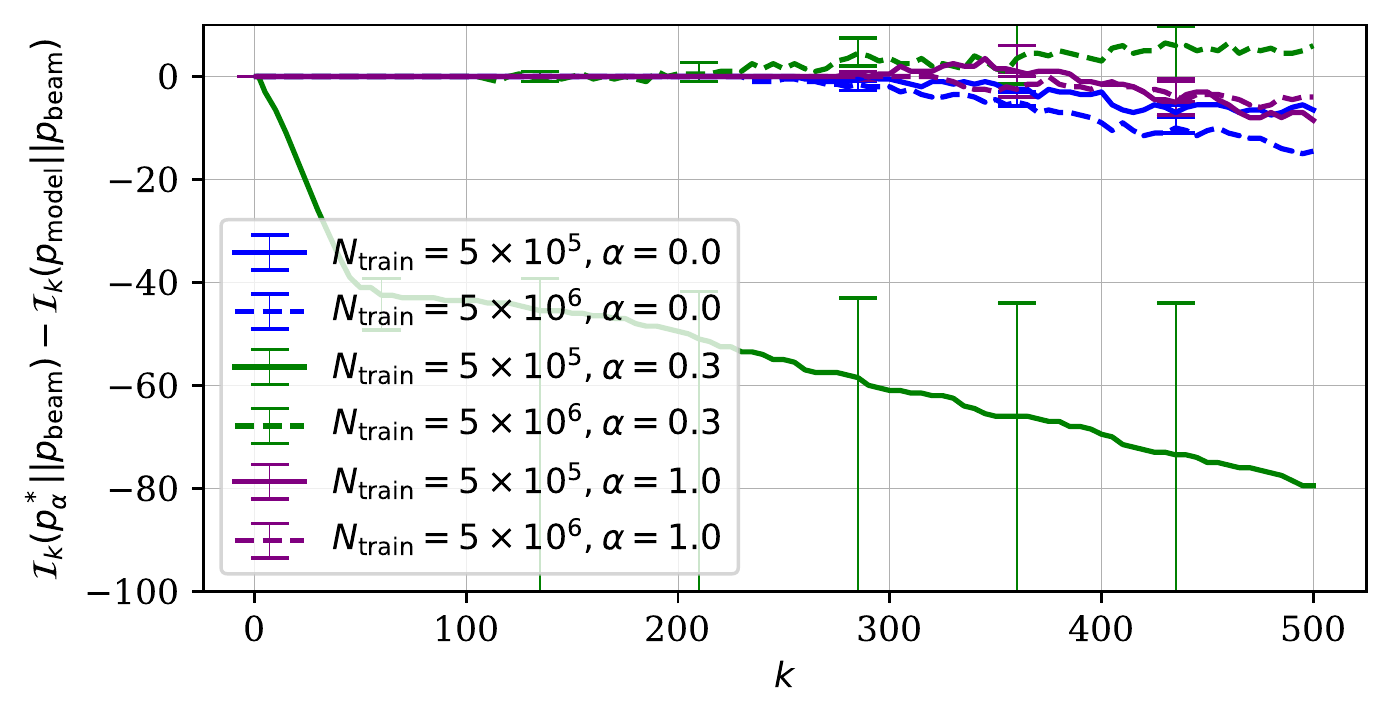}
\end{subfigure}%
\begin{subfigure}{.5\textwidth}
    \includegraphics[width=1.\linewidth]{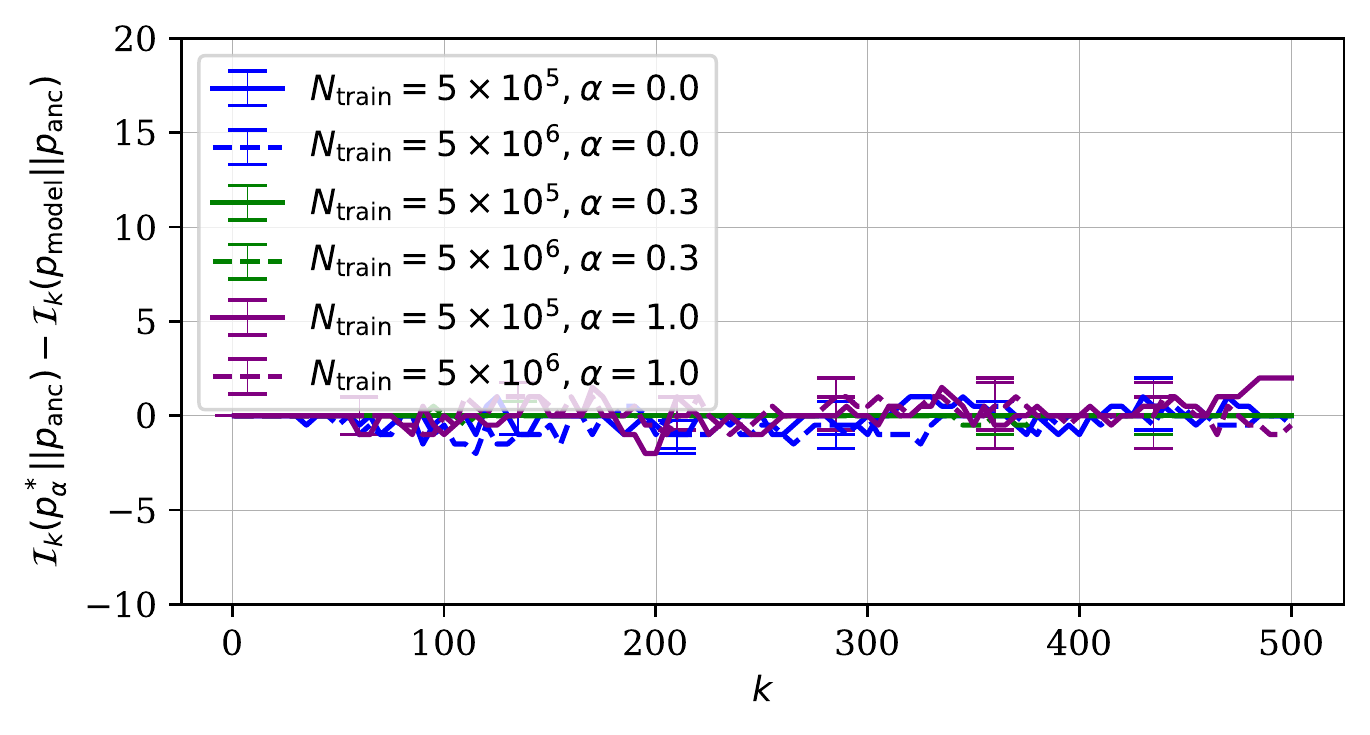}
\end{subfigure}
\caption{$k$-mode set overlap reduction from the ground-truth distribution to the learned distribution using the decoding-induced distribution as a function of $k$. 
The choice of the decoding algorithm affects the pattern of mode recovery degradation along the entire learning chain.
}
\label{fig:kmodeoverlap_reduction_true_model_dec500}
\end{figure*}

First, we study how well the decoding algorithm recovers modes from the learned distribution.
\cref{fig:kmodeoverlap_model} shows $k$-mode set overlap between learned and decoding-induced distributions using both beam search (left) and ancestral sampling (right). Both algorithms fail increasingly more often as $k$ increases. Ancestral sampling fails substantially more often than beam search. This is expected given that ancestral sampling was not designed to find highly probable sequences, unlike beam search. Both of these decoding algorithms fail to recover modes from the learned distribution most when the learned distribution was obtained using the semi-structured ground-truth distribution ($\alpha=0.3$), regardless of the size of the dataset. In other words, the choices made earlier along the learning chain impact the decoding-induced distribution's ability to recover modes from the learned distribution, regardless of which decoding algorithm was used.

Second, we investigate how the choice of the decoding algorithm influences the difference in how the decoding-induced distribution recovers modes of ground-truth and learned distributions.
We thus look at the $k$-mode set overlap reduction from ground-truth to learned distributions ($\mathcal{I}_k(p^*_{\alpha} \| p_\mathcal{F}) - \mathcal{I}_k(p_\text{model} \| p_\mathcal{F})$) for both beam search and ancestral sampling. The positive overlap reduction in \cref{fig:kmodeoverlap_reduction_true_model_dec500} means that the decoding algorithm fails more to recover modes from the learned distribution than from the ground-truth distribution. 

Each decoding algorithm shows a different pattern of the overlap reduction. Reduction is more or less flat and is close to zero for ancestral sampling regardless of the choice of the dataset.
It is, however, different with beam search where we have three observations. First, the reduction overlap deviates from zero as $k$ increases. 
Second, with the semi-structured ground-truth distribution ($\alpha=0.3$) the overlap deviates most, which is then followed by the unstructured variant ($\alpha=0.0$). Third, the number of training points $N_\text{train}$ leads to significant difference in the case of the semi-structured distribution. Reduction overlap goes very negative with the smaller number of training instances, while the trend flips when we have ten times more data. We thereby conclude that the pattern of mode recovery degradation along the entire learning chain depends on the choice of the decoding algorithm.

\section{Conclusion}

In this paper, we studied the propensity of neural autoregressive sequence models to assign high probabilities to sequences that differ from those in the ground-truth distribution.
To measure this phenomenon, we defined \textit{mode recovery cost}, which measures a distribution's ability to recover the highly probable sequences of another distribution. 
We developed a testbed for evaluating mode recovery cost throughout the entire learning chain.

We provided evidence of non-trivial mode recovery cost within this testbed, and observed that the increase in the cost relies heavily on the structuredness of the ground-truth distribution.
Mode recovery from earlier distributions was more costly along the learning chain when the ground-truth distribution was constructed as a product of fully-structured and fully-unstructured distributions such that it reflects patterns in real data.

Mode recovery cost at each stage depended on all the choices made earlier at all the previous stages.
The empirical distribution induced during data collection recovered modes from the ground-truth distribution imperfectly regardless of the dataset size. It was particularly high when we used the semi-structured ground-truth distribution. As expected, mode recovery cost was negatively correlated with a number of training instances.

Mode recovery after learning was directly affected by the choice of the ground-truth distribution as well. 
In general, the learned distribution failed to recover modes from the ground-truth distribution as well as the empirical distribution does. This trend flipped, however, when the learned distribution was parameterized identically to the ground-truth distribution.
Distributions induced during decoding recovered modes of learned distributions with significantly different costs depending on all choices made at previous stages of the learning chain.
The choice of decoding algorithm was also found to influence patterns of mode recovery cost.
Based on these observations, we conclude that we have to use the entire learning chain to study mode recovery in neural autoregressive sequence modeling.

\paragraph{Future directions.} We highlight three main directions of research based on our findings and conclusions. First, mode recovery along the learning chain must be studied in the context of real world problems. To do so, there is a need for future work on approximation schemes of mode recovery cost computable in real tasks. 
Second, the relationship between the ground-truth and learned distributions may be changed to better match real-world cases,  
for instance by considering structured ground-truth distributions that are less similar to the learned model family, or unstructured components that are informed by sequence content.
Third, we have considered standard practices of neural autoregressive 
modeling while constructing the learning chain. 
Extending the learning chain to study the effects of new approaches 
such as knowledge distillation \citep{Kim_2016} or back translation \citep{Sennrich_2016}
is another fruitful direction for future research.

\bibliographystyle{acl_natbib}
\bibliography{anthology,acl2021}

\end{document}